\pdfoutput=1

\documentclass[11pt]{article}

\usepackage{acl}

\usepackage{times}
\usepackage{latexsym}

\usepackage[T1]{fontenc}

\usepackage[utf8]{inputenc}

\usepackage{microtype}

\usepackage{amsmath, amsthm, amssymb, amsfonts}

\usepackage{booktabs}
\usepackage{graphicx}
\usepackage{float}

\usepackage{pifont}
\newcommand{\cmark}{\ding{51}}%
\newcommand{\xmark}{\ding{55}}%

\title{Cross-Lingual QA as a Stepping Stone for\\Monolingual Open QA in Icelandic}

\author{Vésteinn Snæbjarnarson \\
 Miðeind ehf\\
 University of Iceland \\
  \texttt{vesteinn@mideind.is} \\\And
  Hafsteinn Einarsson \\
  University of Iceland\\
  \texttt{hafsteinne@hi.is} \\}

\begin{document}
\maketitle
\begin{abstract}
It can be challenging to build effective open question answering (open QA) systems for languages other than English, mainly due to a lack of labeled data for training. We present a data efficient method to bootstrap such a system for languages other than English. Our approach requires only limited QA resources in the given language, along with machine-translated data, and at least a bilingual language model. To evaluate our approach, we build such a system for the Icelandic language and evaluate performance over trivia style datasets. The corpora used for training are English in origin but machine translated into Icelandic. We train a bilingual Icelandic/English language model to embed English context and Icelandic questions following methodology introduced with DensePhrases (Lee et al., 2021). The resulting system is an open domain cross-lingual QA system between Icelandic and English. Finally, the system is adapted for Icelandic only open QA, demonstrating how it is possible to efficiently create an open QA system with limited access to curated datasets in the language of interest.
\end{abstract}

\section{Introduction}
Open QA systems are question-answering systems that suggest answers to questions by searching through a text corpus. Such systems have improved significantly in recent years, which can, to a large extent, be attributed to transformer-based vector representations of text that are well suited for the task~\cite{vaswaniattention}. The most successful systems have been trained with a focus on English using large datasets such as Natural Questions \cite{kwiatkowski2019natural} (>320k questions), and SQuAD \cite{squad_v2_sv} (>150k questions). In some cases, questions have been generated from text using large generative neural networks \cite{alberti-etal-2019-synthetic}. For most languages, such large datasets do not exist, and the generative models do not perform as well as for English which constitutes the bulk of the training data. For this reason, we investigate what performance can be reached in QA for Icelandic, a language with low QA resources. In that investigation, we study the question of whether English QA data can aid QA system development through the use of machine translation.

In this paper, we present a method to bootstrap an Open QA system for Icelandic where just a few thousand labelled data entries are available. We adapt the DensePhrases~\cite{densephrases} method by applying a bilingual language model, and machine-translated data, in a cross-lingual manner to create a monolingual Open QA system for Icelandic, the first of its kind built exclusively for the language. An overview of the build process is shown step by step in Figure~\ref{fig:hero}.

\section{Related work}
\subsection{Reading comprehension and Open QA}
Open-domain question answering methods look for answers to a given question in a given text corpus (for a recent survey, see~\cite{DBLP:journals/corr/abs-2101-00774}). These methods can be contrasted with reading comprehension (RC) style methods that identify an answer to a question within a single document. The RC methods are useful when an answer is sought in a given text, often referred to as the context. Open QA methods are open in the sense that the questions they can handle are open ended given a large enough underlying corpus. Open QA can be thought of as a generalization of reading comprehension since the answer is typically retrieved from a large collection of text instead of a single document. We note that most open QA methods are extractive, meaning that the suggested answer is found verbatim within a given document. There are also QA methods that provide an answer without explicitly searching through a corpus. For example, the answer can be generated based on knowledge embedded in learned parameters of a system such as GPT-3~\cite{brown2020language}. While promising, the non-extractive methods are not considered in this paper.

Open QA methods solve a common issue in information retrieval where it is not known in what document an answer lies. The simpler reading comprehension methods can be used as components in open QA systems by combining them with a \emph{retriever} component. BM25~\cite{bm25}, a TF-IDF variant, is an example of a commonly used retriever that ranks context based on term frequencies that are shared with the question and their overall commonality. The top documents found by the retriever can then be fed to the reading comprehension component along with the question. The reading comprehension component can, for example, be a fine-tuned variant of a neural language model such as BERT \cite{bertdevlin}. The reading comprehension component is trained to predict the start and end location of an answer span or report whether an answer is not found within the given context by training on a dataset of context and question pairs.

\subsection{Fast retrieval and DensePhrases}
In recent years, efforts in improving open QA have focused on speeding up the lookup of documents, for example, by taking advantage of neural methods. Such a speedup has been realized by embedding documents and questions as dense vector representations such that lookup can be based on fast similarity search where the inner product of the question vector and document vector is used as a proxy for their similarity \cite{dpr, orqa, lin_pretrained_2021}. The embedding function can be trained such that a given question will, with a good chance, lead to the correct document being the highest ranked in the similarity search. The embedding function can also be trained end-to-end by basing the loss function on the performance of looking up the answer. A downside of these methods, in particular the end-to-end systems, is that they can be expensive to train since the document embeddings need to be updated often as a result of updates to the embedding function \cite{Guu2020RetrievalAL}, which can be particularly expensive when many documents need to be embedded repeatedly throughout the training process. Some mitigations have been suggested; as is the case in DensePhrases \cite{densephrases}, which is the foundation of our approach. 

In DensePhrases, segments from documents are first embedded using a phrase model (and fixed), then a query model is trained to embed questions such that the inner product of question embeddings and correct context embeddings are maximized. For an incorrect pairing, the model is trained such that the inner product is minimized instead. 

Fast databases intended for lookup with maximum inner product search (MIPS)~\cite{faiss} enable systems such as DensePhrases to provide answers from massive datasets in subsecond time, making them excellent candidates for production-grade QA systems where an answer and its source can be reported.

\subsection{Multilingual and cross-lingual QA}
In cross-lingual QA, the question and answer are not required to be in the same language, and in multilingual QA the aim is to search for answers in a multilingual corpus. Multilingual QA is not necessarily cross-lingual since the answer can be generated in the same language as the query.

Interest in cross-lingual QA is likely reflected in the growing number of QA datasets in foreign languages~\cite{rogers_qa_2021}. For reading comprehension, it has been shown that multilingual LMs such as mBERT fine-tuned in an English reading comprehension task are capable of zero-shot transfer to other languages such as Japanese, French, and Hindi~\cite{siblini_multilingual_2021, gupta_bert_2020}. Multilingual QA has been performed by extending models for English by using machine translation (MT) on the query and answer~\cite{asai_xor_2021}, MT has also been used to adapt an English semantic parsing model for other languages~\cite{sherborne2020bootstrapping,moradshahi2020localizing}. Multilingual QA was recently implemented without explicit use of MT by extending the Dense Passage Retriever model from~\newcite{dpr} with a fine-tuned mT5 model as an answer generator~\cite{asai2021one}. The answer generator receives top-scoring multilingual passages along with the question and desired answer language to generate the answer. This flexible approach even generalizes to languages not seen in the QA training process thanks to the diverse training set for crosslingual retrieval. A similar approach with an answer generator has also been applied where passage candidates come from different monolingual corpora, and the question is translated and embedded with several monolingual language models~\cite{muller_cross-lingual_2021}.


\subsection{Icelandic QA data}
Currently, a single extractive dataset exists for Icelandic, NQiI~\cite{nqiLREC}. It is a small Icelandic dataset containing only $\sim$5k question-context pairs, half of which have no answer. The dataset is sourced from the Icelandic Wikipedia following the methodology introduced in TyDi-QA~\cite{clark_tydi_2020}. This limited amount of Icelandic QA data is the main reason we translate English QA datasets.

\begin{figure*}
    \centering
  \includegraphics[width=\textwidth]{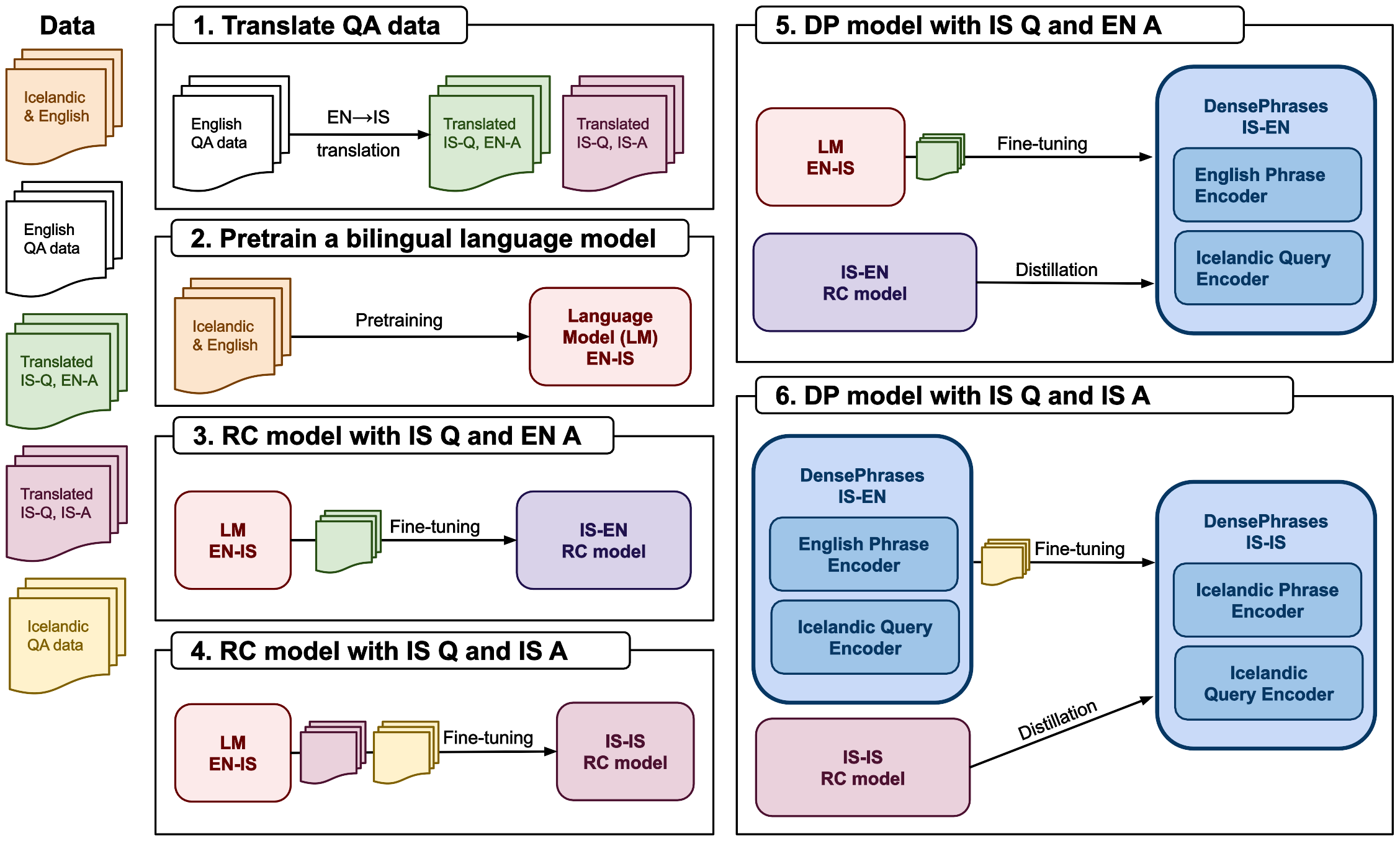}
  \caption{Diagram showing how to train an Icelandic DensePhrases model in six steps.}
  \label{fig:hero}
\end{figure*}

\section{Methods}
\subsection{Translating QA data}
In the first step of the process, we use an English-Icelandic translation system~\cite{wmt2021} for translating NewsQA~\cite{trischler-etal-2017-newsqa}, SQuAD and Natural Questions (NQ)~\cite{kwiatkowski2019natural}. We reviewed the translated questions from SQuAD and out of 100 randomly sampled questions we found that 80 were properly translated such that the meaning was fully preserved.

We translate questions, answers and contexts independently and use a fuzzy matching algorithm (see Appendix~\ref{app:asa}) to map translated answers to spans in the translated context. We refer to the fully translated versions of the datasets as NewsQA-IS, SQuAD-IS, and NQ-IS. For the translated versions of the datasets where only the questions are answered as we use NewsQA-ISQ, SQuAD-ISQ, and NQ-ISQ (for an overview, see Table~\ref{table:qa_data}).

In DensePhrases, questions are generated for all spans of length 0--20 words in the English Wikipedia using a fine-tuned T5~\cite{raffel2020exploring} model. As no such model currently exist that can reliably generate Icelandic, we also translate the generated questions. The spans themselves can not be easily translated as the available models are mostly good at translating well-formed sentences. We refer to this dataset as DP-ISQ. For an overview of all QA datasets used see Table~\ref{table:qa_data}.

\subsection{Pre-training an Icelandic--English language model}
A bilingual language model for Icelandic and English was trained following the base XLM-RoBERTa implementation \cite{xlmr}. We refer to this model as LM EN-IS. The Icelandic training data is the same as the one used for IceBERT \cite{IceBERTLREC}. The Books 3 corpus\footnote{This is similar to \cite{bookcorpus} and was made available in the issue section of the GitHub repository \url{https://github.com/soskek/bookcorpus/issues/27}.} is used as source for English data, it contains around 100GB of data text from a variety of books. The model was trained for 220k updates using a batch size of 8k completing 27 epochs over the data.
%
%
\begin{table*}[htb]
  \centering
  \begin{tabular}{l l c c r }
   \toprule
\textbf{Original dataset} & \textbf{Transl. dataset} & \textbf{Question transl.} & \textbf{Context transl.} & \textbf{Step}\\ \midrule
SQuAD & SQuAD-ISQ & \cmark & \xmark & 3 \\ 
NewsQA & NewsQA-ISQ & \cmark & \xmark & 3 \\
NQ & NQ-ISQ & \cmark & \xmark & 3 \\ \midrule
SQuAD & SQuAD-IS & \cmark & \cmark & 4 \\
NewsQA & NewsQA-IS & \cmark & \cmark & 4 \\
NQ & NQ-IS & \cmark & \cmark & 4 \\ \midrule
DensePhrases (generated) & DP-ISQ & \cmark & \xmark & 5 \\
NQiI & -- & \xmark & \xmark & 6 \\
\bottomrule
  \end{tabular}
    \caption{Overview of QA-datasets used in training and how they were translated. The last column refers to steps where the data is used for fine-tuning in Figure~\ref{fig:hero}.}
\label{table:qa_data}
\end{table*}

\subsubsection{Training RC models}
After pre-training, the bilingual model (LM EN-IS) is fine-tuned for cross-lingual RC where questions are asked in Icelandic and answered in English (step 3 in Figure 1). We fine-tune using SQuAD-ISQ, NewsQA-ISQ, and NQ-ISQ. We refer to this model as the IS-EN RC model. 

The bilingual model (LM EN-IS) is also fine-tuned for an Icelandic only reading comprehension task (step 4 in Figure 1) using the fully translated datasets, NQ-IS, SQuAD-IS and NewsQA-IS along with NQiI. We refer to this model as the IS-IS RC model.

These RC models are later used as a teacher models \cite{hinton2015distilling}. The IS-EN RC model is distilled in the fifth step and the IS-IS RC model in the sixth step of the build process when fine-tuning the Open QA system. Note that to be compatible with the training of the DensePhrases model, these models do not predict missing answers.

\subsubsection{Training cross-lingual DensePhrases}
We also fine-tune the bilingual model (LM EN-IS) to train a DensePhrases setup \footnote{With minor adjustments to work with the SentencePiece \cite{sentencepiece} tokenization used by the bilingual model}. We use the partially translated DP-ISQ dataset to train the cross-lingual DensePhrases model. The result is a phrase encoder that accepts English and a query encoder that accepts Icelandic. Following the DensePhrases approach, we distil the IS-EN RC model at training time. This distillation step can be beneficial since the comparison in the DensePhrases setup is based on an inner product operation, whereas the RC model was trained in a cross-attention setting. This distillation step improved the EM score by 2 points for the original DensePhrases paper and could be validated through ablation in our low-resource setting as well. We refer to the crosslingual DensePhrases model as DensePhrases-IS-EN 

\subsubsection{Training Icelandic only DensePhrases}
In the last step of our process, we take the cross-lingual model DensePhrases-IS-EN and fine-tune it on NQiI to develop a fully Icelandic Open QA system. In this final step, we also distil the IS-IS RC model from the fourth step of the build process. We refer to the final Icelandic only model as DensePhrases-IS.

\section{Results}

\subsection{Reading comprehension model performance}
A comparison of RC performance is shown in \ref{table:en_squad_xlmr_comp}. The table includes performance for the English model RoBERTa and untranslated SQuAD data (for the subset of the data that was successfully translated). Using the bilingual model only leads to a slight drop in performance (-1.7 F1). Translating the data further decreases the performance (-2.7 F1, row 6 in the table) but not catastrophically in any sense. In comparison, fine-tuning on an Icelandic only model (IceBERT) improves performance slightly (+0.6 F1, row 7 in the table). These models are not used in any of the steps shown in Figure~\ref{fig:hero} but the results validate not only the adequacy of the translation method applied, they also demonstrate that the bilingual model is suitable to be adapted for QA in both Icelandic and English. All models were trained for 4 epochs, using a learning rate of 3e-5, maximum sequence length of 512 tokens and a document stride of 128.

\begin{table*}[htb]
  \centering
  \begin{tabular}{c l l l r r}
   \toprule
\textbf{Step} &\textbf{Task} & \textbf{Model} & \textbf{Fine-tuning dataset} &  \textbf{F1} & \textbf{EM} \\ \midrule
- & RC-EN-EN & RoBERTa (EN) & SQuAD & \textbf{75.9} & 74.3 \\
- & RC-EN-EN & LM EN-IS & SQuAD & 74.2 & 73.0 \\\midrule
- & RC-IS-EN & LM EN-IS & NQ-ISQ & 74.9 & 67.1 \\
- & RC-IS-EN & LM EN-IS & SQuAD-ISQ & 59.9 & 50.6 \\
3 & RC-IS-EN & LM EN-IS & NQ-ISQ + SQuAD-ISQ & 75.8 & 67.9 \\ \midrule
- & RC-IS-IS & LM EN-IS & SQuAD-IS & 71.5 & 70.1 \\
- & RC-IS-IS & IceBERT (IS) & SQuAD-IS & \textbf{72.1} & 70.6 \\
4 & RC-IS-IS & LM EN-IS & NewsQA-IS + SQuAD-IS + NQiI & *67.4 & 64.8 \\
\bottomrule
  \end{tabular}
    \caption{Performance in reading comprehension for a mono- and crosslingual setting. RC-X-Y denotes reading comprehension where the question language is X and the answer language is Y. For fine-tuning in the crosslingual setting (RC-IS-EN) in rows 3, 4 and 5, questions have been translated into Icelandic while the context and answers are in English (step 3) whereas the last three rows correspond to fine-tuning on Icelandic only (step 4). The evaluation data in the last row marked with a (*) is from a combination of the datasets used. The best performance on each task is shown in bold.}
\label{table:en_squad_xlmr_comp}
\end{table*}

Performance of the IS-EN RC model is measured on the development set of NQ with translated questions. We fine-tune on NQ-ISQ and SQuAD-ISQ, which refer to the Natural questions and SQuAD datasets with only the questions machine translated into Icelandic (step 3, row 5 in the table). Another RC model was fine-tuned on fully translated QA data along with NQiI (step 4, row 8 in the table). We chose that model for use in the fourth step since it was trained on more data than the models in rows 6 and 7 with a small sacrifice in performance on SQuAD-IS, 70.80 F1 and 69.51 EM. With $\sim$2/3 questions answered exactly, we conclude that the RC models serve well as a teacher models for the DensePhrases training (steps 5 and 6).

\subsection{Open QA performance}
Performance for the cross-lingual Open QA system (DensePhrases-IS-EN, from step 5) is shown in Table \ref{table:dph open} where results are evaluated for the Natural Questions test-dataset, both for the version with machine-translated questions (Is--En) and the original one (En--En). The system still performs well on the English only data. For reference, we note that the original DensePhrases model~\cite{densephrases} had an exact match score of 40.9 on NQ and 39.4 on SQuAD when the query-side encoder was fine-tuned for those datasets, respectively.

\begin{table*}[htb]
  \centering
  \begin{tabular}{c l l l r r r r}
    \toprule
    \textbf{Step} & \textbf{Task} & \textbf{Method} & \textbf{Data}  & \textbf{EM} & \textbf{F1} & \textbf{EM top 10}& \textbf{F1 top 10} \\ \midrule
5 & Open QA IS-EN & XL-DensePhr. & NQ-ISQ & 11.3 & 15.2 & 29.6 & 38.5 \\
5 & Open QA EN-EN & XL-DensePhr. & NQ & 14.0 & 18.9 & 34.7 & 45.0 \\ \midrule
6 & Open QA IS-IS & XL-DensePhr. & NQiI & 9.7 & 18.8 & 26.8 & 44.6 \\
6 & Open QA IS-IS & XL-DensePhr. & G.betur & 6.0 & 8.3 & 14.8 & 20.6\\
6 & Open QA IS-IS & XL-DensePhr. & Trivia & 5.4 & 6.9 & 14.6 & 18.4\\
- & Open QA IS-IS & BM25 + IB-QA & NQiI & 2.4 & 17.9 & 2.4 & 18.1  \\
- & Open QA IS-IS & CORA & NQiI & 15.0 & \textbf{28.6} & - & -\\\bottomrule

\end{tabular}
  \caption{Performance for open QA in a cross-lingual Icelandic and English (DensePhrases-IS-EN) setting and in a monolingual IS-IS setting (DensePhrases-IS). In the cross-lingual setting, the performance on NQ is included for reference. All the models are based on the bilingual model (LM EN-IS) except for the second to last one, which corresponds to using the IceBERT model along with BM25, and the last one, which is based on mT5. We highlight in bold the best performance in Open QA on the NQiI dataset.}
\label{table:dph open}
\end{table*}

The Icelandic open QA system (DensePhrases-IS, from step 6) is evaluated on NQiI as well as datasets suitable for open QA in Icelandic, the Gettu betur corpus (4,569 questions with answer) \cite{IceQA} and Icelandic Trivia Questions\footnote{Available online at \url{https://github.com/sveinn-steinarsson/is-trivia-questions}} (11,610 questions with answers). We note that these datasets are not guaranteed to contain answers that are present in the Icelandic Wikipedia, but serve as a future baseline for Open QA in Icelandic.

Performance results for the model in the sixth step are shown in Table \ref{table:dph open}. For comparison, a BM25 + IceBERT-QA result is included. The results are not as good as reported for English systems in, e.g. \cite{dpr}, which we currently attribute to the small size of the NQiI dataset.

Finally, we embed the Icelandic Wikipedia for use with CORA~\cite{asai2021one} using the models released with the paper. The NQiI test dataset is used for evaluation. This method significantly outperforms the one presented in this paper as shown in the last row of Table~\ref{table:dph open} with F1 28.6 and EM 15.0.

\section{Discussion and future work}

As noted in the literature review, good results have been achieved in multilingual QA using an answer generator to generate an answer in a selected language~\cite{asai2021one}. For a monolingual setting, our approach provides a way to create an Open QA system without an answer generator as in the original DensePhrases approach.

The model used in the original DensePhrases is SpanBERT \cite{spanbert} whereas we trained a bilingual RoBERTa \cite{roberta} model that has been proven to be successful for Icelandic. For future work, a bilingual SpanBERT model is likely to improve performance as reported in the original paper.

We also evaluated the CORA method on NQiI and it surpassed our method by a significant margin, highlighting the value of training models in a multilingual manner and using a generative model. CORA was not trained specifically on Icelandic QA although it is based on mT5 which was pre-trained on corpora that includes some Icelandic. The result highlights the potential of crosslingual transfer for QA in low-resource languages.

Finally, we emphasize that the quality of the resulting model of the process presented in this paper is affected by multiple factors. For example, it is related to the performance of the translation method but possibly also to language intricacies. A greater amount of training data for Icelandic QA, along with human translated pairs of questions and contexts would cast of light of the penalty incurred from using MT data. We believe the results can be much better with a larger and higher quality target language QA dataset, noting that, e.g. the answer span labelling in the NQiI is somewhat inconsistent. However, we also believe that QA for Icelandic is challenging, and we encourage others to try it out.

\section{Conclusion}
We have shown how to build an Open QA system from scratch for Icelandic, a language with very limited original QA resources. We first develop a cross-lingual QA system by taking advantage of English QA-data, a well performing translation model, a bilingual language model and the DensePhrases approach. This system is then adapted for monolingual Open QA. The method is not perfect but shows some promising results.

\section*{Acknowledgements}
We thank Prof. Dr.-Ing. Morris Riedel and his team for providing access to the DEEP super-computer at Forschungszentrum Jülich. We also thank the Icelandic Language Technology Program~\cite{nikulasdottir_language_2020}, it has enabled the first author to focus on work in Icelandic NLP. Finally, we thank the anonymous reviewers for their helpful feedback.

\bibliography{anthology,custom}
\bibliographystyle{acl_natbib}

\appendix
\section{Answer span alignment}\label{app:asa}
We apply a heuristic matching method to align the translated questions with spans in the translated context. The method does not rely on more complex word alignment methods between the source text and the translated text but is based on translating the answer and looking up the translated answer in the translated answer context using a fuzzy Levenshtein distance.

In our matching method, we search for the translated answer and then the original answer in the translated context. If either is found, we label the matched string as the answer. Otherwise, we apply a fuzzy matching approach. Denote by $w_t$ the number of words in the translated answer. We perform a sliding window search over all contiguous sequences of words in the translated context that contain $w_t$, $w_t-1$, $w_t+1$ many words. We label and return a sequence as the answer in the translated setting if the Levenshtein distance between the translated answer and the sequence exceeds 0.9. If no sequence is sufficiently similar to the translated answer, we repeat this sliding window search using the original answer instead of the translated answer. If neither search was successful, we would discard the translated question-context pair from training in the fourth step.

Only 6,893 questions, 4.8\% of the total data, were discarded from the SQuAD dataset using the matching method since an answer span could not be labelled. 11,478 questions, 9.6\% of the total, were discarded from the NewsQA dataset. The only publicly released reading-comprehension style Icelandic dataset for QA, Natural Questions in Icelandic (NQiI) \cite{nqii}, is also used for training.

\end{document}